\documentclass[a4paper,fleqn]{cas-dc}

\usepackage{graphicx}
\usepackage{cite}
\usepackage{algorithm, algorithmic}
\usepackage[authoryear,longnamesfirst]{natbib}
\usepackage{natbib}
\usepackage{stfloats}

\setcitestyle{numbers,square}

\usepackage{longtable}
\usepackage[normalem]{ulem}
\useunder{\uline}{\ul}{}

\begin{document}
\let\WriteBookmarks\relax
\def\floatpagepagefraction{1}
\def\textpagefraction{.001}

\title[mode = title]{ HAT-GAE: Self-Supervised Graph Auto-encoders with Hierarchical Adaptive Masking and Trainable Corruption}

\tnotemark[1]

\author[1]{Chengyu Sun}
\author[1]{Liang Hu}
\author[1]{Hongtu Li}
\author[1]{Shuai Li}
\author[1]{Tuohang Li}
\author[1]{Ling Chi\corref{cor1}}
\address[1]{College of Computer Science and Technology, Jilin University, Changchun, 130000, China}
\cortext[cor1]{Corresponding author} 
\ead{E-mail addresses:cysun20@mails.jlu.edu.cn (C. Sun),hul@jlu.edu.cn (L. Hu),lihongtu@jlu.edu.cn (H. Li), shuaili18@mails.jlu.edu.cn (S. Li),lith19@mails.jlu.edu.cn (T. Li), chiling@jlu.edu.cn.}

\begin{abstract}[S U M M A R Y]
Self-supervised auto-encoders have emerged as a successful framework for representation learning in computer vision and natural language processing in recent years, However, their application to graph data has been met with limited performance due to the non-Euclidean and complex structure of graphs in comparison to images or text, as well as the limitations of conventional auto-encoder architectures. In this paper, we investigate factors impacting the performance of auto-encoders on graph data and propose a novel auto-encoder model for graph representation learning. Our model incorporates a hierarchical adaptive masking mechanism to incrementally increase the difficulty of training in order to mimic the process of human cognitive learning, and a trainable corruption scheme to enhance the robustness of learned representations. Through extensive experimentation on ten benchmark datasets, we demonstrate the superiority of our proposed method over state-of-the-art graph representation learning models.

\end{abstract}
\begin{keywords}
graph representation learning \sep generative learning \sep graph auto-encoder \sep self-supervised learning
\end{keywords}
\maketitle

\section{Inroduction}\label{Indro}
Graph data, which is a powerful tool for understanding and analyzing the relationships between various entities, is widely used in a variety of real-world applications\cite{du2022disentangled}, such as social networks \cite{bourigault2014learning}, biological networks \cite{theocharidis2009network}, and domain-specific multimedia data. Graph data is usually composed of nodes and edges, where each node represents a single data point, and each edge reflects the relationship between two nodes. For instance, in a social network, each node  represent an individual and the edges between them depict relationships, such as classmates or coworkers.

Graph representation learning, which seeks to convert raw graph data into a representation vector that preserves intrinsic graph properties, is crucial for effectively analyzing and understanding the relationships and patterns within the graph\cite{mo2022simple,li2022transo}. For example, in the social network described above, we can predict the valuable traits of a person, such as their purchase intentions, by examining the behavior of people with similar relationships\cite{wang2022causal}. Inspired by the success of Recurrent Neural Network(RNN)\cite{zaremba2014recurrent} and convolutional neural network(CNN)\cite{lecun1998gradient} in Natural language processing(NLP) and computer vision(CV), the primary efforts of graph representation learning focused on primarily supervised training paradigms such as Graph Convolutional Networks (GCN)\cite{kipf2016semi}, Spectral Graph Convolutional Networks(SGCN)\cite{derr2018signed} and Graph Attention Networks(GAT) \cite{velivckovic2017graph}. However, the inadequacy of manually labeled data has long been a weakness of supervised learning, resulting in inefficiency in label-sensitive scenarios\cite{zhang2022unsupervised}. As a result, self-supervised learning (SSL), which can extract informative knowledge without relying on manual labeling, has emerged as a practical and appealing learning paradigm for graph data.

Based on the training strategy, self-supervised learning methods can be generally divided into two categories: contrastive learning (CL) and generative learning (GL). 

Contrastive learning models are grounded in the idea of maximizing mutual information (MI)\cite{hjelm2018learning}, which involves training the model to predict the agreement between two augmented graphs. GL, on the other hand, can be traced back to auto-encoders \cite{hinton2006reducing}, which are trained to compress data features into low-dimensional representations using an encoder network and then attempt to reconstruct the input vectors using a decoder network. Despite generally outperforming generative learning models in terms of performance, contrastive learning has limitations and weaknesses. One contributing factor to the effectiveness of CL methods is their heavy reliance on complex training strategies\cite{hou2022graphmae,feng2022adversarial}. For instance, the use of bi-encoders with momentum update and exponential moving average can be essential for stabilizing the training of GCC\cite{qiu2020gcc} and BGRL\cite{thakoor2021large}. Moreover, many contrastive methods, such as DGI \cite{velickovic2019deep}, GRACE\cite{zhu2020deep}, and GCA \cite{zhu2021graph}, require negative samples obtained through laborious sampling from graphs. In addition, CCA-SSG\cite{zhang2021canonical} is prone to hyper-parameters due to its reliance on sophisticated data augmentation strategies primarily based on heuristics\cite{feng2022adversarial,lee2022augmentation,yu2022graph}.

On the other hand, generative learning models, which aim to directly reconstruct the input graph data without requiring additional complex precautions, can naturally avoid the inherent issues in contrastive models. For illustration, consider GAE \cite{kipf2016variational} as an example. This method utilizes a GNN-based encoder to generate node embeddings from the input graph, and a decoder to reconstruct the adjacency matrix from these embeddings. To further improve efficiency, VGAE\cite{kipf2016variational} combines the GAE method with the concept of a variational auto-encoder\cite{kingma2013auto}in order to achieve graph representation learning goals. There are several variations of GAE/VGAE developed with the aim of increasing performance. MGAE\cite{2017MGAE} aims to recover the raw features of the input graph from corrupted, noisy features. Graph Completion \cite{2020When} focuses on predicting masked node features from the features of neighboring nodes. AttrMasking\cite{hu2019strategies} not only reconstructs node attributes, but also edge attributes. GATE\cite{salehi2019graph} uses both feature and link reconstruction to learn graph representations. GALA \cite{2020Symmetric} reconstructs the original feature matrix by training a Laplacian smoothing-sharpening graph auto-encoder model. GPT-GNN \cite{hu2020gpt} presents an autoregressive framework for iteratively reconstructing both nodes and edges. SuperGAT\cite{kim2022find}reconstructs the adjacency matrix from the latent representations of every layer in the encoder.

Despite the successful development of graph GL methods, data reconstructing schemes which are proved to be a critical component for graph representation learning \cite{hou2022graphmae}, have yet to receive much attention in the existing literature. We argue that an effective data reconstructing schemes should satisfy three fundamental properties: Moderate complexity of reconstruction strategy,Adaptivity of corruption scheme, and Hierarchy of model training.

\textbf{Moderate complexity of reconstruction strategy.} In contrast to traditional techniques used in computer vision, defining a reconstruction strategy in generative learning is non-trivial due to the complex, non-Euclidean nature of graphs\cite{li2022curvature}. We argue that an effective reconstruction strategy should have a moderate level of complexity, as both too low and too high complexity can negatively impact the model to extract informative graph features. For instance, GAE only reconstructs embeddings from the original graph, which is straightforward but may not provide sufficient informative gradient for the model to optimize its objective. On the other hand, GATE reconstructs both nodes and edges to learn the representation, which may result in an over-complex objective that hinders the model from learning valuable embeddings.

\textbf{Adaptivity of corruption scheme.} Corrupting the original input graph is an essential step in generative learning, and it is common practice in the existing literature to disturb the input graph randomly \cite{hu2019strategies,hu2020gpt,2017MGAE}. For example, in MGAE, the model tries to reconstruct the raw features from corrupted features processed by random noise. However, this can result in the loss of important information and guide the model in the wrong direction. Instead, the corruption strategy should be adapted to the graph, such as corrupting sub-critical dimensions of the feature while preserving the valuable ones for the model to learn from.

\textbf{Hierarchy of model training.} Similar to human cognitive learning habits, it is more beneficial for skill development to progressively increase the difficulty of learning from easy to hard. Applying this learning strategy to the training of a model may further improve its performance. However, most previous works\cite{hu2019strategies,hu2020gpt,2017MGAE,salehi2019graph,kipf2016variational,kim2022find,kingma2013auto,kim2022find,2020Symmetric} do not focus on designing a difficulty gradient for the model and instead give it a fixed task to train from scratch. This can result in low performance in the early stages of training due to the excessive difficulty of learning. On the other hand, if we can set a hierarchical difficulty gradient for the model and start the initial training at the lowest difficulty, incrementally  increasing the difficulty as training progresses, it may lead to the model extracting more valuable information.

\begin{figure*}[ht]   \centering  \includegraphics[height=8cm, width=17cm]{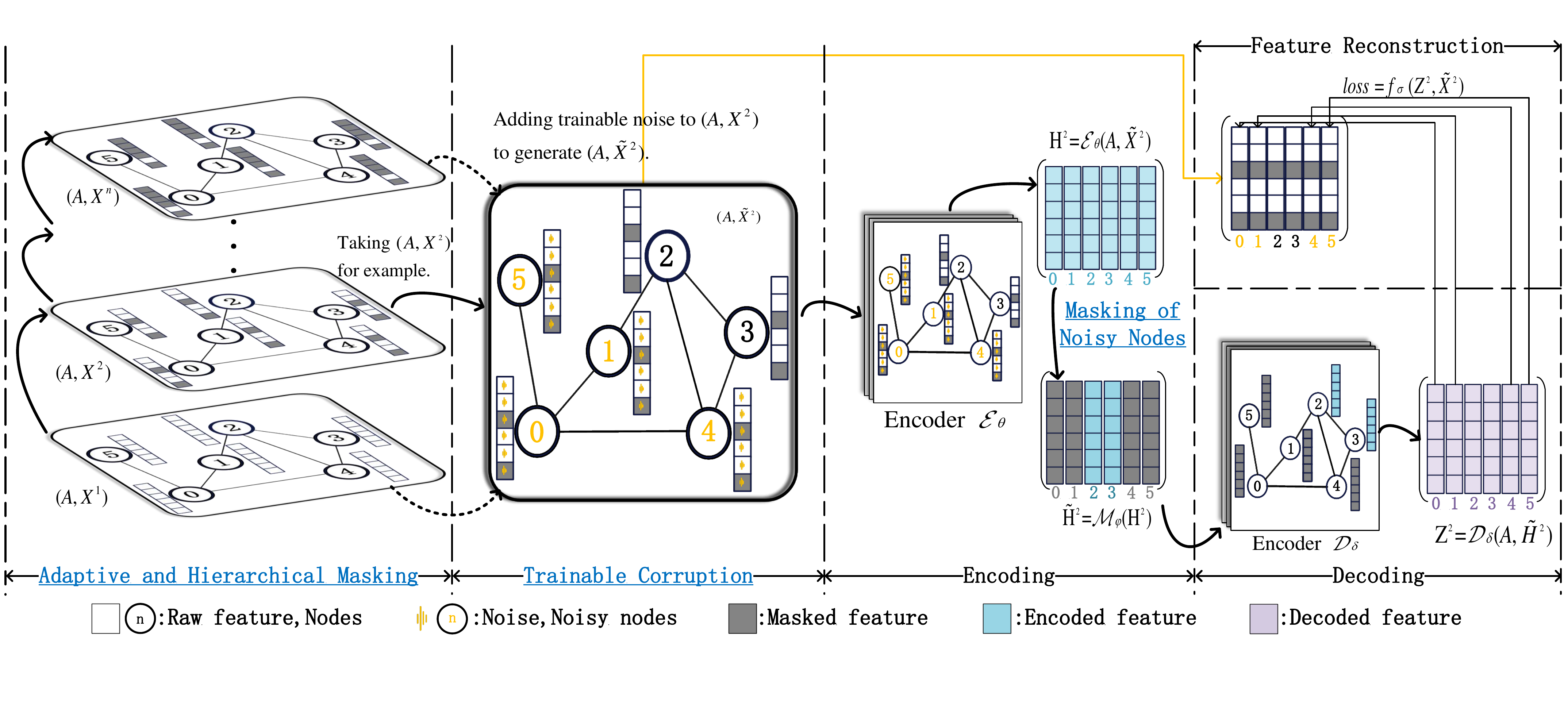}  \caption{Our proposed Self-Supervised Graph Auto-encoders with Hierarchical Adaptive Masking and Trainable Corruption(HAT-GAE) model.} \label{HAT-GAE}  
\end{figure*}

To meet the three properties outlined above, we present a novel generative framework for unsupervised graph representation learning, referred to as Self-Supervised Graph Auto-encoders with Hierarchical Adaptive Masking and Trainable Corruption (HAT-GAE for short). As illustrate in Figure \ref{HAT-GAE},HAT-GAE employs an adaptive masking technique to selectively preserve important dimensions of the node features in the input graph, while incrementally increasing the number of masks in order to improve the difficulty of learning over time. Then, HAT-GAE introduces trainable noise that is learned during training to the features of certain nodes in the graph. The noisy graph is then fed into an encode-decoder framework, in which the features of the noisy nodes are zeroed out before being passed to the decoder to further enhance the model's learning challenge. Finally, HAT-GAE uses the task of feature reconstruction to encourage the extraction of expressive embeddings.

\textbf{Our contribution}

\begin{itemize}
\item[$\bullet$]Firstly, We propose a novel generative framework for unsupervised graph representation learning,referred to as Self-Supervised Graph Auto-encoders with Hierarchical Adaptive Masking and Trainable Corruption (HAT-GAE)  that fulfills the three essential properties discussed above. Compared to prior models, HAT-GAE has the ability to adaptively corrupt graph data, automatically increase the learning difficulty and learn the corruption strategy during training, resulting in enhanced representational power of the model's output(section \ref{Methodology}).

\item[$\bullet$]Secondly, we conduct comprehensive empirical studies using ten public benchmark datasets of different scale and categories on node classification under transductive and inductive experiment settings. The result show that HAT-GAE consistently outperforms state-of-the-art methods and even surpasses its contrastive counterparts, demonstrating its great potential in real-world applications(section \ref{The result and analysis}).
\end{itemize}

\section{Related work}
Our work is related to the following three topics:

\textbf{Self-Supervised graph representation learning} Self-supervised learning gained traction in graph representation learning due to its ability to extract informative representations \cite{velickovic2019deep,hassani2020contrastive} through well-designed training strategies, without the need for supervised signals. Graph representation learning aims to learn semantic and structural information from a graph and generate informative embeddings that can be used as input features for downstream tasks such as node classification, graph classification, and clustering. Given the expressive nature of graph structures, most effective methods for graph representation learning are based on graph neural networks (GNNs), which use recursive message passing to learn complex dependencies within the graph. There are several types of GNNs, including graph convolutional networks (GCN) \cite{welling2016semi}, graph isomorphism networks (GIN) \cite{xu2018powerful}, and graph attention networks (GAT) \cite{velickovic2017graph}, among others.

\textbf{Contrastive learning models} With the revival of the classical principle of mutual information (MI), previous works are extensively explored contrastive learning patterns in computer vision. In recent years, researchers adopted similar contrastive frameworks to enable self-supervised training on graph data. However, existing graph contrastive methods often require the ability to distinguish positive and negative data pairs from a large number of examples \cite{he2020momentum}, which can be time-consuming. For example, Deep Graph InfoMax \cite{velickovic2019deep}, the earliest work in this area, randomly shuffles node features and uses an MI-based loss to discriminate between positive/negative samples. Based on DGI, GRACE, GCA \cite{zhu2021graph}, and GraphCL leverage in-batch negative samples, GCC \cite{qiu2020gcc} uses a negative sampling principle similar to MoCo to compare a node to the contextual information of the node. MVGRL\cite {hassani2020contrastive} generates negative samples through graph diffusion \cite{klicpera2019diffusion}. Although negative sampling is not necessary for BGRL, a sophisticated training strategy such as momentum updates and the exponential moving average is essential for stable training. Additionally, all of these contrastive methods require a carefully designed augmentation strategy, which can be challenging to define in a principled way.

\textbf{Auto-encoder-based graph generative learning models} Auto-encoder is trained to compress the initial high-dimensional data vectors into a low-dimensional representation using an encoder network and then attempts to reconstruct the initial data vectors using a decoder network. On the other hand, the generative model is trained by reconstructing the corrupted input graph and using the graph itself as supervision signals. In order to improve performance, many efforts in recent years combined the powerful representational capabilities of auto-encoder and the practical training paradigm of generative model for graph representation learning. The earliest works GAE\cite{kipf2016variational} employs a decoder based on the inner production function to reconstruct the adjacency matrix of the input graph from the encoding matrix encoded by a GCN-based encoder. VGAE incorporates the idea of variational auto-encoder into GAE. In pursuit of more informative graph representation, SIG-VAE\cite{hasanzadeh2019semi} extends GAE by integrating the idea of variational inference. ARGA/ARVGA\cite{pan2018adversarially} adopt the paradigm of GAE/VGAE to generative adversarial networks (GANs). SuperGAT\cite{kim2022find} recovers the raw feature from the latent representations of every layer in the encoder. In addition to the models trained by rebuilding an input graph's structure information(adjacency matrix), there are important works to obtain graph representation by reconstructing the feature information(node feature). For example, Graph Completion\cite{2020When} is trying to recover the masked node features from the information of neighboring nodes. MGAE\cite{2017MGAE} attempts to reconstruct the raw features from corrupted features processed by random noise. GALA\cite{2020Symmetric} rebuilds the feature matrix by training a Laplacian smoothing-sharpening graph auto-encoder model. 

In recent years, there has been significant progress in auto-encoder-based generative learning. However, the performance of generative models still falls short compared to that of contrastive learning counterparts. This paper aims to identify the weaknesses of existing generative models and design a model that can match or surpass the performance of contrastive models on the node classification task.

\section{Methodology}\label{Methodology}
\subsection{Problem Statement}
Formally, We denote a graph as ${{\cal G}}{\rm{ = (}}A{\rm{, }}X{\rm{)}}$. The feature matrix and adjacency matrix of ${{\cal G}}$ are represented by $X\in{\mathbb{R}^{N \times F}}$ and $A\in{\{ 0,1\} ^{N \times N}}$,respectively .$V = \{ v_1,v_2,...,v_N\} $ and $E\in V\times V$ denote the node set and edge set of ${{\cal G}}$,respectively. $x_i\in{\mathbb{R}^F}$ is the feature of node $v_i$,where $x_{iu}$ represent the u-th dimension of node i, and $A_{ij} = 1$ if and only if  $(v_i,v_j) \in E$.

Our proposed HAT-GAE aims to follow a self-supervised paradigm to train a GNN-based encoder ${{\cal E}}_\theta \left( {{\cal G}} \right)$ and generate informative node embeddings for downstream tasks such as node classification.

\subsection{Our Proposed Framework}
As shown in Figure \ref{HAT-GAE}, we first use Adaptive and Hierarchical Masking to corrupt the input graph ${{\cal G}} = (A,{X^1})$ to create a hierarchical graph set $\{ (A,{X^i})\;|\;i = 2,3,...,n\} $, each of which will be fed iteratively to the next module, i.e., Trainable corruption. Take $(A,{X^2})$ for example, we add trainable noise to the portion of nodes in the graph to generate $(A,{\tilde X^2})$ . Then, we feed $(A,{\tilde X^2})$ to a GNN-based encoder ${\cal E}_{\theta}$ to obtain a hidden feature matrix ${H^2}$ , followed by a masking operation on the noisy nodes to generate the masked hidden matrix ${\tilde H^2}$. After that, we input ${\tilde H^2}$ to a GNN-based decoder ${\cal D}_{\delta}$ to obtain the encoded representation ${Z^2}$. Finally, we train the model by reconstructing the ${Z^2}$ to ${\tilde X^2}$ ,where only the noisy nodes are involved. 

Our model consist of five components: Adaptive and Hierarchical Masking, Trainable Corruption, Encoding, Decoding, and Feature Reconstruction. In the following sections, we will provide a detailed description of our model.

\subsubsection{Hierarchical Adaptive Masking}
We propose a novel graph coruption strategy, referred to as Adaptive and Hierarchical Corruption, which incorporates two layers of considerations.

\textbf{Adaptability of the strategy.} Most previous work \cite{manessi2021graph,park2019symmetric,zhu2020self} rely on uncontrolled random corruption on features of nodes, such as random feature masking, which can lead to the loss of hard connections that contain essential information and result in the drop of critical gradient during model training. However, the proposed adaptive corruption strategy can overcome these drawbacks by adaptively filtering out sub-important information and preserving critical information for the model to learn, leading to improved performance.

Intuitively, We assume that feature dimensions that frequently appear in influential nodes are critical. For example, in a citation network where nodes represent papers and each feature dimension corresponds to a keyword, the keywords that frequently appear in highly influential papers should be considered informative and important. Therefore, measuring the importance score of a dimension starts with measuring the importance score of a node.

\textbf{Measuring Nodes Importance.} For convenience of description, let us take ${{\cal G}} = (A,{X^1})$ as an example. we evaluate the important score $S_v$ with in-degree of node v,defined as:

\begin{equation}
	S_v =in-de(v) = \sum\limits_{u \in V}^{} {{A}_{uv}} 
\end{equation}
where ${A}_{uv}$ is the value in the u-th row and v-th column in the adjacent matrix  ${A}$. To provide some intuition of the measurement of in-degree, take the example mentioned above; in a citation network, a paper with more connections pointing to it, which suggests that the paper has been widely referenced and may be influential in its field, should be considered crucial.  

In addition to the in-degree, other widely-used measures for evaluating the importance score of a node include eigenvector centrality\cite{bonacich1987power,muller1995neural} and PageRank\cite{muller1995neural,page1999pagerank}.  These methods are based on different principles and provide different perspectives on the importance of a node in a network.

Eigenvector centrality assigns relative scores to all nodes in a graph based on the principle that connections to high-scoring nodes contribute more to the score of the node than equal connections to low-scoring nodes. The eigenvector centrality $S_v$ of node $v$ defined by:
\begin{equation}
	S_v = {1 \over \lambda }\sum\limits_{u \in N(v)} {s_u}  = {1 \over \lambda }\sum\limits_{u \in V} {{A}_{vu}s_u} 
\end{equation}
where $N(v)$ is the set of neighbors of node $v$ and $\lambda$ is a  largest eigenvalue of adjacent matrix $A$. 

PageRank is a variant of eigenvector centrality that evaluates the importance of nodes in a graph based on the principle that a node with many incoming links is more important than a node with fewer incoming links. The PageRank score for a node $v$, denoted as $s_v$, is defined as:

\begin{equation}
{S_v} = \alpha {A}{D^{ - 1}}s + (1 - \alpha ){e \over n}
\end{equation}
where $A$ is the adjacency matrix of the input graph  ${{\cal G}} = (A,{X^1})$, $D$ is the degree matrix with the diagonal element as the degree of each node, $D^-1$ is the inverted degree matrix, $e$ is a vector of all ones and $n$ is the number of nodes in the graph. The parameter $\alpha$, known as the damping factor, is a value between 0 and 1 that controls the probability of the random surfer following the links the links on the page and  $(1 - \alpha ){e \over n}$ is the teleportation vector that ensures that the random surfer may jump to any node with probability $(1 - \alpha )$ if it falls into a sink.

While eigenvector centrality and PageRank offer more sophisticated ways of evaluating the importance of a node, our experimental results show that using in-degree as a measure of node importance is a simple yet effective and computationally efficient option in our proposed model.

\textbf{Measuring Dimensions Importance.} After calculating the importance scores for each node, we proceed to calculate the importance scores for each dimension in the feature vector. The importance score $Sd_{u}$ in the u-th dimension is calculated as the weighted sum of the of the importance scores of all nodes, where the weight is determined by the absolute value of the u-th dimension of the feature in the graph. It is defined mathematically as follows:

\begin{equation}
Sd_u\; = \sum\limits_{v \in V}^{} {s_v} \; \times |{X^1}_{vu}|\;\;
\end{equation}
where $|{X^1}_{vu}|\;$ denote the absolute value of node $v$’s feature in dimension u.  

Then, we generate a set that consist of all calculated importance score of all dimensions $Sd = \{ Sd_1,Sd_2,...,Sd_F\} $, where $F$ is the number of dimensions in feature vector. We sort the set in descending order according to the values to obtain a sorted set $\tilde Sd$, where the further back an item is in the set $\tilde Sd$, the more important the dimension it corresponds to. Next, we sample $F \times pf$ values from the front of the sorted set and apply an adaptive masking function $m_{pf}(x_i)$ to mask the corresponding dimensions of the feature vectors for all nodes.The process is represented as follows:

\begin{equation}\label{adaptive masking}
{X^2} = [m_{pf}({X^1}_1),m_{pf}({X^1}_2),...,m_{pf}({X^1}_N)]
\end{equation}
where the hyper-parameter  $pf$ controls the percentage of dimensions that will be masked, $N$ is the total number of nodes in the graph, and $X_2$ is the generated feature matrix.

\textbf{Hierarchical Masking of the strategy.} As illustrated in Figure \ref{HAT-GAE}, we iteratively apply the equation \ref{adaptive masking} to the graph data generated from the previous iteration at regular intervals, defined by the hyper-parameter $num$ that controls the number of round of hierarchical masking applied \label{rounds}. For example, if we train the model for $2000$ epochs and set $num$ to $200$, the graph will be masked every $200$ epochs, for a total of $2000/200 = 10$ masking operations.The hierarchical masking procedure is as following:   

\begin{equation}\label{Hierarchical masking}
{X^n} = [m_{pf}({X^{n - 1}}_1),m_{pf}({X^{n - 1}}_2),...,m_{pf}({X^{n - 1}}_N)]
\end{equation}

After n iterations,the obtained $X^n$  and the adjacent matrix $A$ of input graph constitute a hierarchical graph set $S = \{ ({X^n},A)|n = 1,2,3...\} $. We input the graph set to the next component sequentially. 

Please note that, in order to prevent all feature dimensions from being set to zero, causing the gradient to vanish, we incrementally  mask feature dimensions across multiple iterations by applying the hyper-parameter $pf$ recursively to the number of remaining feature dimensions from the previous iteration. Let's say the feature vector dimension for a node is $F$ and $pf$ is the masking rate, at each iteration i, the number of masked feature dimensions(m) is defined as:

\begin{equation}\label{internal}
{m_i}{\rm{ }} = {\rm{ }}\left( {F - \left( {i - 1} \right)*{m_{i - 1}}} \right)*pf,i{\rm{ }} > {\rm{ }}1
\end{equation}

In the first iteration i=1, the formula will be ${m_1}{\rm{ }} = {\rm{ }}d*pf$.

At the beginning of training, the number of masked feature dimensions is limited, leaving more valid knowledge for the model to learn from, resulting in a lower training difficulty. As the training iterations progress, more feature dimensions are masked, providing fewer learnable knowledge for the model, and increasing the training difficulty. This process aligns with the cognitive learning process of human and our experiments shown that it leads to significant improvements in the model's performance.

\subsubsection{Trainable Corruption}

As previously discussed in last section, the hierarchical graph set $S$ is processed in an iterative manner and passed on to the Trainable Corruption module where a trainable noise is introduced to a selected subset $\tilde V \in V$. Formally,it is achieved by sampling a random matrix $M \in {\left\{ {{{({0^F})}^T},{\rm{ (}}{1^F}{)^T}} \right\}^N}$ with dimensions $N*F$, wherein each element is independently drawn from a Bernoulli distribution $M_i\sim\beta (1 - pn)$, with $pn$ as the hyper-parameter that regulates the corruption rate for all nodes and is referred to as \textsl{the noisy rate}. Subsequently, the generated node feature ${\tilde X^{\rm{n}}}$ is computed by:

\begin{equation}
{\tilde X^n} = {X^n} + W_n \odot {M^T}\;,\;\;n = 1,2,3,...
\end{equation}
where $W_n$ represents the trainable noise and addition operation following it applies the trainable noise to the designated subset $\tilde V \in V$. Then ,we sequentially input the generated hierarchical set $\tilde S = \{ ({\tilde X^n},A)|n = 1,2,3...\} $ into the next module.

\subsubsection{Encoding}
We utilize a multi-layer graph attention network with four attention heads as our encoder ${{\cal E}}_\theta $ to transform the hierarchical graph set $\tilde S = \{ ({\tilde X^n},A)|n = 1,2,3...\} $ into respective hidden codes $\{ {H^n} \in {\mathbb{R}^{N \times F'}}|n = 1,2,3...\}$. The hidden state $h_v$ corresponding to node $v$ is computed as:

\begin{equation}
h_v = \mathop {||}\limits_{k = 1}^k \sigma (\sum\limits_{u \in NB(v)} {{a^k}_{vu}} \;\;{W^k}h_u)
\end{equation}
where $NB(v)$ denotes the neighbour node of $v$, $k$ represents the number of attention heads, $\sigma $ is a activation function. The attention coefficient from node $u$ to $v$, $a_{vu}$, is defined as:

\begin{equation}
a_{vu} = \;{{\exp ({\rm{LeakReLU}}({\alpha ^T}[Wh_v||Wh_u]))} \over {\sum\nolimits_{k \in NB(v)} {\exp ({\rm{LeakReLU}}({\alpha ^T}[Wh_v||Wh_k]))} }}
\end{equation}
where LeakReLu is a non-linear function and $\alpha  \in {\mathbb{R}^{2F'}}$ is a trainable vector.

To further increase the challenge of model training, we set the hidden code corresponding to the noisy node to zero, as follows:

\begin{equation}
{\tilde H^n} = M\varphi ({H^n}) = {H^n} \odot {M^T},n = 1,2,3,...
\end{equation}

\subsubsection{Decoding and Feature Reconstruction}
We employ a GAT decoder ${{\cal D}}_\delta $ with same structure to the encoder to maps the hidden code ${\tilde H^n}$ to the final embedding:

\begin{equation}
{{\rm{Z}}^{\rm{n}}}{\rm{ = }}{{\cal D}}_\delta {\rm{(}}A{\rm{, }}{\tilde H^{\rm{n}}}{\rm{)}}
\end{equation}

We compute the cosine similarity to evaluate the distance between the corresponding node and reconstruct the noisy feature ${\tilde X^n}$ from generated embedding $Z^n$,defined as:

\begin{equation}
dst(x_i,x_j) = {{{{x_i}^T}x_j} \over {||x_i|| \cdot ||x_j||}}
\end{equation}

The final loss is defined as:

\begin{equation}
\ell  = \;{1 \over {|\tilde V|}}\sum\limits_{v \in \tilde V} {{{(1 - dst(\;{{\tilde X}_v}^n,{Z_v}^n))}^2}} 
\end{equation}
where ${\tilde X_v}^n$ and ${Z_v}^n$ are corrupted feature and encoded hidden code corresponding to node $v$, respectively.

\begin{table*}[t]
	\begin{tabular}{cccccc}
		\hline
		Dataset & \#Nodes & \#Edges & \#Features & \#Class & Category \\ \hline
		Cora & 2,708 & 5,429 & 1,433 & 7 & citation network \\
		Citeseer & 3,327 & 4,732 & 3,703 & 6 & citation network \\
		Pubmed & 19,717 & 44,338 & 500 & 3 & citation network \\
		Amazon-Photo & 7,650 & 119,081 & 745 & 8 & co-purchase network \\
		Amazon-Computer & 13,752 & 245,861 & 767 & 10 & co-purchase network \\
		Coauthor-CS & 18,333 & 81,894 & 6,805 & 15 & academic networks \\
		Coauthor-Physics & 34,493 & 247,962 & 8,415 & 5 & academic networks \\
		Ogbn-arxiv & 169,343 & 1,166,243 & 128 & 40 & citation network \\
		PPI & 56,944 & 818,716 & 50 & 121 & protein network \\
		Reddit & 231,443 & 11,606,919 & 602 & 41 & social network \\ \hline
	\end{tabular}
	\caption{The statistics of the datasets that are used to evaluate the performance of our model}
	\label{datasets_table}
\end{table*}

\section{EXPERIMENTS}
\subsection{Experimental Setup}
In this section, we conduct empirical evaluations of our proposed model on node classification using ten publicly available benchmark datasets. We design experiments to investigate the following research questions:
\begin{itemize}
\item[$\bullet$]RQ1: Does our model exhibit superior versatility, i.e. can it adapt to different types and sizes of datasets?
\item[$\bullet$]RQ2: Does our model outperform state-of-the-art methods?
\item[$\bullet$]RQ3: Do the proposed hierarchical adaptive masking and trainable corruption schemes contribute to the performance of the proposed model? How does each module impact model performance? (Ablation Studies)
\item[$\bullet$]RQ4: Is the proposed model sensitive to hyperparameters? How do key hyperparameters impact model performance?(Sensitivity Analysis)
\end{itemize}

\subsubsection{Datasets}
For a comprehensive comparison, we use seven widely-used datasets to evaluate the performance of our model on transductive node classification and three large-scale datasets to study the performance on inductive node classification. The statistics of the datasets are summarized in Table \ref{datasets_table}.

\begin{itemize}
\item[$\bullet$]Cora, Citeseer, Pubmed \cite{sen2008collective} and Ogbn-arxiv \cite{hu2020open} are citation networks, where nodes represent papers and edges indicate citation relationships.The label of a node corresponds to a article category.
\item[$\bullet$]Amazon-Computers and Amazon-Photo \cite{shchur2018pitfalls} are networks of co-purchase relationships constructed from Amazon, where nodes represent goods and edges connect goods that are frequently bought together. Each node is labeled indicating its category.
\item[$\bullet$]Coauthor-CS and Coauthor-Physics \cite{shchur2018pitfalls} are two academic networks that contain co-authorship graphs based on the Microsoft Academic Graph from the KDD Cup 2016 challenge. In these graphs, nodes represent authors and edges indicate co-authorship relationships, meaning two nodes are connected if they have co-authored a paper. The label of an author corresponds to their most active research field.
\item[$\bullet$]PPI \cite{zitnik2017predicting} is a biological protein-protein interaction network that contains multiple graphs, with each graph corresponding to a human tissue, where each node has multiple labels that are a subset of the gene ontology set.
\item[$\bullet$]Reddit \cite{hamilton2017inductive} is a large-scale social network that contains Reddit posts belonging to different communities (subreddit). In the dataset, nodes correspond to posts and edges connect posts if the same user has commented on both.
\end{itemize}

\subsubsection{Evaluation protocol}
For every experiment, we follow the linear evaluation scheme as introduced in \cite{velickovic2019deep}, where each model is firstly trained in an unsupervised manner. Then, we freeze the parameters of the encoder and generate embeddings for all the nodes. After that, the resulting embeddings generated by encoder are used to train and test a simple $\ell_2$-regularized logistic regression classifier. To ensure fairness, we train the model for twenty runs and report the averaged performance on each dataset. Furthermore, we measure performance using micro-averaged F1-score on inductive tasks and accuracy on transductive tasks. Please note that for inductive learning tasks, tests are conducted on unseen or untrained nodes and graphs, while for transductive learning tasks, we use the features of all data, but the labels of the test set are masked during training. We follow the public data splits \cite{hassani2020contrastive,velickovic2019deep,zhang2021canonical} of Cora, Citeseer, and PubMed. Regarding the other seven datasets, since they have no public splits available, we instead randomly split the datasets, where 10\%, 10\%, and the rest 80\% of nodes are selected for the training, validation, and test set, respectively. 

\begin{table*}[ht]
	\resizebox{\textwidth}{!}{
		\begin{tabular}{@{}cccccccccc@{}}
			\toprule
			\multicolumn{10}{c}{Transductive Task(Accuracy)} \\ \midrule
			& Datasets & Cora & Citeseer & PubMed & Am.Photos & Am.Computers & CoauthorCS & CoauthorPhy & Ogbn-Arxiv \\ \midrule
			\multirow{6}{*}{\begin{tabular}[c]{@{}c@{}}Contrastive\\  Learning\end{tabular}} & DGI & 82.35±0.37 & 72.12±0.29 & 76.70±0.11 & 91.59±0.17 & 83.99±0.30 & 92.21±0.29 & 94.43±0.26 & 70.54±0.20 \\
			& GRACE & 82.05±0.25 & 71.44±0.41 & 80.32±0.33 & 82.21±0.22 & 87.51±0.12 & 92.90±0.07 & 95.21±0.10 & 71.49±0.31 \\
			& MVGRL & 83.44±0.11 & {\ul 73.45±0.05} & 80.22±0.31 & 91.77±0.14 & 87.52±0.24 & 92.07±0.19 & 95.44±0.27 & 70.25±0.12 \\
			& BGRL & 82.69±0.17 & 71.60±0.22 & 79.81±0.09 & 92.65±0.31 & \textbf{89.39±0.11} & {\ul 93.10±0.22} & {\ul 95.49±0.31} & 71.50±0.17 \\
			& InfoGCL & 83.60±0.11 & 73.29±0.26 & 79.24±0.18 & 91.17±0.22 & 87.41±0.34 & 91.89±0.24 & 94.01±0.25 & 69.08±0.20 \\
			& CCA-SSG & 84.12±0.19 & 73.22±0.20 & 78.98±0.35 & 92.10±0.13 & 86.96±0.28 & 92.05±0.19 & 93.86±0.10 & 71.30±0.41 \\ \midrule
			\multirow{5}{*}{\begin{tabular}[c]{@{}c@{}}Generative \\ Learning\end{tabular}} & GAE & 71.33±0.31 & 65.82±0.11 & 72.34±0.27 & 85.27±0.11 & 80.20±0.21 & 90.05±0.35 & 90.89±0.24 & 64.08±0.22 \\
			& GPT-GNN & 80.29±0.10 & 68.51±0.33 & 76.29±0.26 & 89.27±0.28 & 83.36±0.15 & 92.71±0.09 & 93.27±0.16 & 67.14±0.31 \\
			& GATE & 83.25±0.19 & 71.90±0.22 & 81.01±0.31 & 91.91±0.14 & 86.71±0.29 & 92.04±0.11 & 93.52±0.41 & 68.72±0.12 \\
			& GraphMAE & {\ul 84.19±0.35} & 73.41±0.41 & {\ul 81.21±0.37} & {\ul 93.01±0.26} & 88.32±0.16 & 92.79±0.10 & 95.30±0.32 & {\ul 71.59±0.29} \\
			& \textbf{\begin{tabular}[c]{@{}c@{}}HAT-GAE\\ (Ours)\end{tabular}} & \textbf{84.78±0.11} & \textbf{74.28±0.22} & \textbf{81.88±0.14} & \textbf{93.58±0.24} & {\ul 88.55±0.18} & \textbf{93.17±0.25} & \textbf{95.57±0.21} & \textbf{71.99±0.15} \\ \midrule
			\multirow{2}{*}{\begin{tabular}[c]{@{}c@{}}Supervised \\ Learning\end{tabular}} & GCN & 81.81±0.12 & 70.55±0.17 & 78.79±0.14 & 77.19±0.21 & 86.55±0.31 & 92.42±0.34 & 95.50±0.24 & 71.70±0.14 \\
			& GAT & 82.27±0.31 & 72.76±0.14 & 79.45±0.36 & 92.76±0.17 & 86.99±0.27 & 92.17±0.36 & 95.51±0.16 & 71.91±0.19 \\ \bottomrule
		\end{tabular}
	}
	\caption{ Summary of performance on transductive tasks in terms of accuracy in percentage with standard deviation. The highest performance of unsupervised models is highlighted in boldface, and the second-highest performance of unsupervised models is underlined.}
	\label{result_transductive}
\end{table*}

\begin{table*}[ht]
	\resizebox{\textwidth}{!}{
		\begin{tabular}{@{}cccccccccccccc@{}}
			\toprule
			\multicolumn{14}{c}{Inductive Task(F1-score)} \\ \midrule
			& \multicolumn{6}{c|}{Contrastive Learning} & \multicolumn{5}{c|}{Generative Learning} & \multicolumn{2}{c}{Supervised Learning} \\ \midrule
			Baselines & DGI & GRACE & MVGRL & BGRL & InfoGCL & \multicolumn{1}{c|}{CCA-SSG} & GAE & GPT-GNN & GATE & GraphMAE & \multicolumn{1}{c|}{\textbf{HAT-GAE(Ours)}} & GCN & GAT \\ \midrule
			Reddit & 93.91±0.14 & 95.01±0.36 & 94.21±0.10 & 94.31±0.23 & 93.56±0.33 & \multicolumn{1}{c|}{95.11±0.18} & 90.27±0.22 & 93.51±0.20 & 95.10±0.31 & {\ul 95.89±0.24} & \multicolumn{1}{c|}{\textbf{96.06±0.10}} & 95.14±0.11 & 95.95±0.21 \\
			PPI & 63.52±0.24 & 69.59±0.15 & 68.50±0.21 & 73.52±0.16 & 73.14±0.10 & \multicolumn{1}{c|}{73.21±0.23} & 67.51±0.35 & 72.76±0.29 & 73.25±0.18 & {\ul 74.39±0.19} & \multicolumn{1}{c|}{\textbf{74.72±0.28}} & 75.65±0.25 & 97.32±0.32 \\ \bottomrule
		\end{tabular}
	}
	\caption{Summary of performance on inductive tasks in terms of F1-scores in percentage with standard deviation. The highest performance of unsupervised models is highlighted in boldface, and the second-highest performance of unsupervised models is underlined.}
	\label{result-inductive}
\end{table*}

\subsubsection{Baselines}
We evaluate our proposed model against representative baseline methods from two categories: (1) contrastive learning methods, including DGI \cite{velickovic2019deep}, GRACE \cite{zhu2020deep}, MVGRL \cite{hassani2020contrastive}, BGRL \cite{thakoor2021large}, InfoGCL \cite{xu2021infogcl}, CCA-SSG \cite{zhang2021canonical} and (2) generative learning methods, including GAE \cite{kipf2016variational}, GPT-GNN \cite{hu2020gpt}, GATE \cite{salehi2019graph}, GraphMAE \cite{hou2022graphmae}. To directly compare our proposed method with supervised counterparts, we also report the performance of two representative models GCN \cite{welling2016semi} and GAT \cite{velickovic2017graph}, where they are trained in an end-to-end fashion. For all baselines, we report their performance based on their official implementations.

\subsection{The result and analysis}\label{The result and analysis}

\subsubsection{RQ1: Does our model exhibit superior versatility, i.e. can it adapt to different types and sizes of datasets?}
In order to demonstrate the versatility and applicability of our proposed model, we evaluate its performance on a variety of datasets, spanning a range of sizes (from 2708 to 231,443 nodes) and data types (citation networks, co-purchase networks, academic networks, protein networks, social networks). The results of our evaluation are presented in Table \ref{result_transductive} and Table \ref{result-inductive}, where it can be observed that our proposed model demonstrates strong performance across all datasets. This suggests that the model's ability to extract universal information and adapt to different graph data, thereby highlighting the transferability of the proposed method. These findings illustrate the superior versatility of the proposed model, indicating its potential for a wide range of real-world applications.

\begin{table*}[H]
	\begin{tabular}{@{}llllllll@{}}
		\toprule
		Variants & Cora & Citeseer & PubMed & Am.Photos & Am.Computers & CoauthorCS & CoauthorPhy \\ \midrule
		HAT-GAE-AM & 82.12±0.10 & 73.01±0.20 & 79.24±0.31 & 90.39±0.20 & 87.41±0.03 & 91.20±0.14 & 93.28±0.03 \\
		HAT-GAE-HM & 83.21±0.13 & 73.10±0.22 & 80.01±0.25 & 92.65±0.11 & 87.67±0.16 & 91.28±0.25 & 93.95±0.17 \\
		HAT-GAE-TC & 84.01±0.22 & 73.96±0.05 & 81.26±0.04 & 92.92±0.18 & 87.69±0.26 & 92.95±0.05 & 94.69±0.15 \\
		HAT-GAE(full) & \textbf{84.78±0.11} & \textbf{74.28±0.22} & \textbf{81.88±0.14} & \textbf{93.20±0.24} & \textbf{88.55±0.18} & \textbf{93.17±0.25} & \textbf{95.57±0.21} \\ \bottomrule
	\end{tabular}
	\caption{Results of Variants of HAT-GAE on Node Classification Tasks.}
	\label{result-Variants}
\end{table*}

\subsubsection{RQ2: Does our model outperform state-of-the-art methods?}
From the results presented in Table \ref{result_transductive} and Table \ref{result-inductive}, it is evident that in nine out of the ten datasets, our proposed HAT-GAE model outperforms existing unsupervised baseline methods by considerable margins in both transductive and inductive tasks. Furthermore, on the  Coauthor and the Reddit dataset, we observe that while existing baselines already obtained high performance, our method HAT-GAE still pushes the boundary forward. Additionally, it is worth noting that HAT-GAE is competitive with the models trained with labeled supervision on all eight transductive datasets and the inductive dataset Reddit. The strong performance demonstrates the superiority of our proposed generative learning framework. Additionally, we make other observations as follows:

First, while contrastive methods have been shown to achieve superior performance compared to generative counterparts in recent years, our proposed generative learning-based model, HAT-GAE, outperforms contrastive methods in node classification tasks. It is noteworthy that contrastive learning methods tend to rely heavily on complex training strategies, time-consuming negative sampling, and data augmentation techniques, which can make them intricate to design and implement effectively. In contrast, our proposed generative learning-based model is simpler in design and easier to implement, making it more accessible to researchers and practitioners. This is the key advantages of our proposed method over existing contrastive methods.

Second, GPT-GNN is based on an autoregressive framework, which decomposes joint probability distributions as a product of conditionals to perform node and edge reconstruction iteratively. However, the autoregressive framework relies on the assumption of sequential data. Instead of natural language or images, most graphs do not possess inherent ordering. Therefore, autoregressive methods may be less well-suited for graph data, resulting in suboptimal performance compared to other methods that are better tailored to the graph data structure.

Third, it can be observed that GAE and GATE perform poorly on both transductive and inductive settings. One reason for this is that GAE only employs a simple binary link classification task during training, which may not be sufficient to learn higher-level knowledge. Additionally, GATE uses a combination of feature and link reconstruction to learn representation, which can lead to an overly complex training goal and hinder the model from extracting useful embeddings. However, the proposed hierarchical adaptive masking and trainable corruption in our model can incrementally  increase the difficulty of training as it progresses, allowing the model to learn more valuable information.

\subsubsection{RQ3: Do the proposed hierarchical adaptive masking and trainable corruption schemes contribute to the performance of the proposed model? How does each module impact model performance? (Ablation Studies)}

In this section,we design three variant models to study the impact of each critical component of HAT-GAE, named HAT-GAE-AM, HAT-GAE-HM, HAT-GAE-TC, respectively. For HAT-GAE-AM, we substitute adaptive masking component with random masking, where each dimension is masked with a hyper-parameter probability $p_r$, without the adaptive selecting operation. To construct HAT-GAE-HM, we only employ adaptive masking once with equation \ref{Hierarchical masking} before adding trainable noise. Please note that, to comprehensively evaluate the two variant models HAT-GAE-AM and HAT-GAE-HM, we test the all possible hyper-parameters for $p_r$ and $p_f$ respectively and report the best performance, where $p_f$ is adaptive masking rate in equation \ref{adaptive masking}. For HAT-GAE-TC, we remove the operation of trainable corruption, i.e., we input the adaptive and hierarchical masked graph set directly to the encoder. We evaluate the three variant model with seven public real datasets, and the results are presented in Table \ref{result-Variants}.From the results, we can see that the adaptive masking, hierarchical masking, and trainable corruption scheme improve model performance consistently across all datasets. For example, HAT-GAE achieves 2.66\%, 1.57\%, and 0.77\% absolute improvement compared to the other three variant models on cora dataset, respectively. These results verify the effectiveness of our proposed hierarchical adaptive masking and trainable corruption schemes.

\subsubsection{RQ4: Is the proposed model sensitive to hyper-parameters? How do key hyperparameters impact model performance?(Sensitivity Analysis)}
In this section, we perform sensitivity analysis on critical hyper-parameters of HAT-GAE:the adaptive masking rate $p_f$, noisy rate $p_n$, and the number of hierarchical masking internal $num$ as mentioned in section \ref{rounds}.

\begin{figure*}[h]   \centering  \includegraphics[height=6cm, width=14cm]{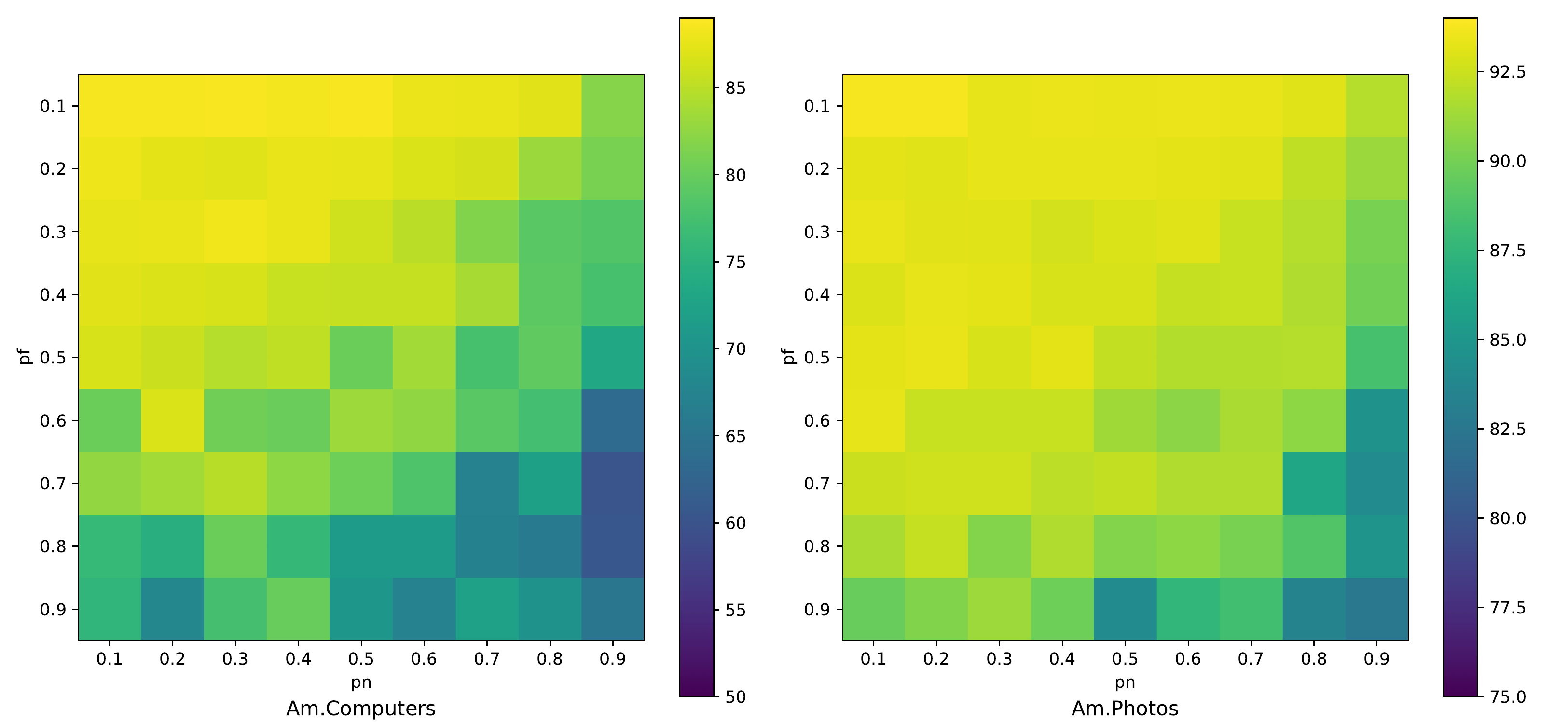}  \caption{Sensitivity Analysis of Hyperparameters in HAT-GAE on $p_f$ and $p_n$.} \label{pf_pn_all}  
\end{figure*}

\textbf{Effect of $p_f$ and $p_n$.} To evaluate the effect of $p_f$ and $p_n$, we train HAT-GAE for 1000 epochs on the Am.Computers and Am.Photos datasets and set the values of $p_n$ and $p_f$ to range from 0.1 to 0.9. The results under different combinations of $p_f$ and $p_n$ are presented in Figure \ref{pf_pn_all}. We observe that the performance of node classification in terms of Micro-F1 is relatively stable when the parameters are not set too large. Thus, we conclude that overall, our model is insensitive to these probabilities, demonstrating its robustness to hyper-parameter tuning. However, if the probability of trainable corruption is set too high (e.g., $pn/pf$> 0.7), the performance can be greatly affected. For example, when $p_n$ = 0.8, the features of the graph are over-corrupted by noise, causing the encoder to extract less true information from the graph and be more affected by the redundant information. This leads the encoder to provide misguided information to the decoder, resulting in low accuracy of node classification.

\begin{figure}[h] 
	   \centering  \includegraphics[height=5cm, width=7cm]{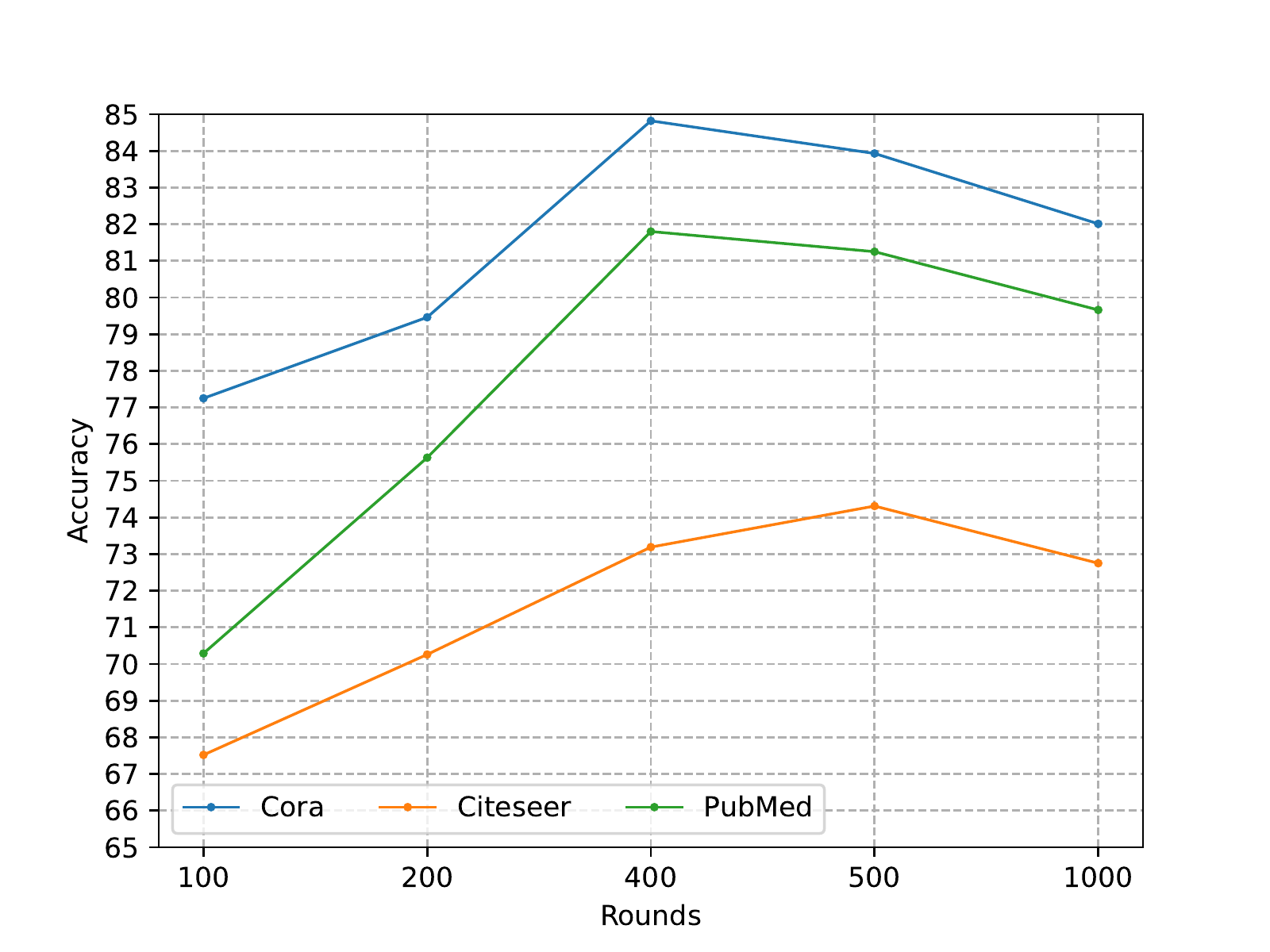}  \caption{The effect of the number of hierarchical masking internal $num$.} \label{pic_hirar} 
\end{figure}

\textbf{Effect of the number of hierarchical masking rounds $num$.}  In order to evaluate the effect of the number of hierarchical masking internal $num$ on model performance, we conducted experiments on the Cora, Citeseer, and Pubmed datasets for 2000 training epochs and varied the number of rounds from 100 to 1000. Specifically, for a given number of rounds, the model applies adaptive masking operations at  $2000/num$ rounds during training, with the number of masked feature dimensions increasing according to equation \ref{internal} for each operation. We report the average performance across 10 runs for each parameter setting and present the results in Figure \ref{pic_hirar}. From the figure, it can be observed that a moderate number of rounds, such as 400 or 500, yields the highest performance. Conversely, a small number of rounds, such as 100 or 200, results in sub-optimal performance, potentially due to the high frequency of masking causing excessive loss of information and hindering the ability of the model to learn useful representations.

\section{Conclusion}
In this work, we presents a novel auto-encoder-based generative model for graph representation learning, which addresses the limitations of conventional auto-encoders in handling the non-Euclidean nature and complex structure of graph data. Our proposed model, which incorporates a hierarchical adaptive masking mechanism and a trainable corruption scheme, is able to incrementally improve the difficulty of training and robustness of learned representations. Through extensive experimentation on various benchmark datasets, we demonstrate the effectiveness of our proposed method and its superiority over state-of-the-art graph representation learning approaches. In addition, this research contributes to a better understanding of the factors affecting the performance of auto-encoders on graph data and presents a promising direction for future work in this field.

\section{Future Directions}
In the future, there are several potential avenues for extending this research. One direction could be to evaluate the proposed model on other applications, such as recommendation systems. Additionally, it would be intriguing to examine the application of the hierarchical adaptive masking and trainable corruption scheme in other generative models for graph representation learning, including variational auto-encoders and generative adversarial networks. Furthermore, incorporating other forms of graph data, such as attributed graphs, multi-modal graphs, or temporal graphs, could also be an interesting area of exploration.

\clearpage

\appendix
\section{Implementation Details}
\subsection{Computing Infrastructures}
We implement our proposed model using DGL 0.8.2 with CUDA 10.2 and PyTorch 1.9.1\cite{fey2019fast}. All datasets used in our experiments are sourced from DGL libraries. Our experiments are conducted on a Linux server equipped with two NVIDIA Tesla V100 GPUs (32GB memory each) and seventy-two Intel Xeon Gold 6240 CPUs.

\subsection{Hyper-parameter Specifications}
In our model, We utilize the Adam optimizer with an initial learning rate of 0.001, utilizing a learning rate decay schedule without warmup. Our model employs a PReLU non-linear activation function and a hidden state dimension of 256$ \sim$1024. The encoder and decoder architectures are same, with two layers of GAT and four attention heads. More specific details about the datasets and hyper-parameters used in our experiments can be found in Table \ref{hiper}.

\begin{table*}[]\label{hiper}
	\resizebox{\textwidth}{!}{
\begin{tabular}{@{}ccccccccccc@{}}
	\toprule
	Hyper-parameters & Cora & Citeseer & PubMed & Am.Photos & Am.Computers & CoauthorCS & CoauthorPhy & Ogbn-Arxiv & Reddit & PPI \\ \midrule
	adaptive masking rate & 0.1 & 0.1 & 0.2 & 0.2 & 0.1 & 0.1 & 0.1 & 0.1 & 0.2 & 0.2 \\
	noisy node rate & 0.5 & 0.5 & 0.5 & 0.5 & 0.6 & 0.5 & 0.5 & 0.5 & 0.75 & 0.5 \\
	internal of hierarchical masking & 200 & 100 & 300 & 300 & 300 & 300 & 300 & 300 & 300 & 300 \\
	hidden size & 512 & 512 & 1024 & 512 & 512 & 512 & 512 & 1024 & 512 & 1024 \\
	max epoch & 500 & 1000 & 1500 & 1500 & 1500 & 1500 & 1500 & 2000 & 800 & 1400 \\
	wight decay & 2e-4 & 2e-5 & 1e-5 & 2e-4 & 2e-4 & 2e-4 & 2e-4 & 0 & 2e-4 & 0 \\ \bottomrule
\end{tabular}
}
	\caption{The hyper-parameters used in our experiments.}
\end{table*}

\bibliographystyle{elsarticle-num}
\bibliography{myrefs}

\end{document}